\documentclass[preprint,12pt,authoryear]{elsarticle}

\usepackage{amssymb}
\usepackage{mathtools}
\usepackage{mathrsfs}
\usepackage{graphicx}
\usepackage{subcaption}
\usepackage[space]{grffile}
\usepackage{url}
\usepackage{booktabs}

\usepackage[pagewise, displaymath, mathlines]{lineno}

\usepackage[colorlinks]{hyperref}
\hypersetup{
  citecolor = {blue}
}

\usepackage{geometry}
\date{}

\begin{document}

\begin{frontmatter}

\title{Reward-Conditioned Attention: How Reward Design Shapes What Autonomous Driving Agents See}

\author[ensia]{Mohamed Benabdelouahad}
\ead{mohamed.benabdelouahad@ensia.edu.dz}

\author[ensia]{Ahmed Djalal Hacini}
\ead{ahmed.hacini@ensia.edu.dz}

\author[uge]{Nadir Farhi}
\ead{nadir.farhi@univ-eiffel.fr}

\author[ensia]{Aissa Boulmerka}
\ead{aissa.boulmerka@ensia.edu.dz}

\address[ensia]{National School of Artificial Intelligence (ENSIA), Sidi Abdellah Campus, Algiers, Algeria.}

\address[uge]{Cosys-Grettia, Univ Gustave Eiffel, F-77454 Marne-la-Vallee, France.}

\begin{abstract}
We investigate how reward design shapes the internal attention patterns of reinforcement learning agents trained for autonomous driving. Using three Perceiver-based agents that share identical architectures and training data but differ only in their reward configurations---ranging from basic violation penalties to continuous proximity penalties---we analyze cross-attention allocation across 50 real-world scenarios from the Waymo Open Motion Dataset. A central methodological finding is that na\"ive pooling of timesteps across episodes substantially underestimates the attention--risk relationship; within-episode correlation with Fisher z-transform aggregation is the appropriate statistic and reveals a robustly positive link between collision risk and agent-directed attention. Building on this validated methodology, we demonstrate two reward-conditioned effects: agents trained with navigation rewards allocate up to $2.0\times$ more attention to GPS-path tokens than those trained with additional proximity penalties---and $4.7\times$ more than agents with no navigation incentive---revealing that reward content directly determines which scene elements the encoder prioritizes, and continuous time-to-collision penalties create a \textit{learned vigilance prior}---elevated resting agent surveillance maintained throughout collision-free phases. In several scenarios, the complete-reward and minimal-reward models exhibit opposite attention--risk correlation directions, demonstrating that reward design can qualitatively reverse attentional strategy rather than merely modulating its magnitude. These results suggest that attention analysis is a practical diagnostic for verifying that a reward function produces the intended representational behaviour in safety-critical RL systems.
\end{abstract}


\end{frontmatter}

\section{Introduction}
\label{sec:intro}

Reinforcement learning (RL) has demonstrated strong potential for autonomous driving (AD), with recent work achieving expert-level accuracy in closed-loop simulation \citep{charraut2025vmax, cusumano2025robust}. However, deploying RL-trained driving policies requires understanding \textit{what} these agents have learned to attend to and \textit{why}. While the ``attention is not explanation'' debate \citep{jain2019attention, wiegreffe2019attention} has established that attention weights are not necessarily faithful indicators of model reasoning, a distinct and complementary question remains underexplored: does the \textit{reward function} used during training predictably shape what the encoder attends to?

This question matters for two reasons. First, if reward design systematically influences attention patterns, practitioners can use attention analysis as a diagnostic tool to verify that their reward function teaches the agent to monitor the right aspects of the driving scene. Second, understanding how different reward components shape internal representations informs principled reward engineering---a central challenge in RL for AD \citep{kiran2022deep}.

We study this question in the V-Max framework \citep{charraut2025vmax}, which provides transformer-based driving agents trained with Soft Actor-Critic (SAC) \citep{haarnoja2018sac} on the Waymo Open Motion Dataset (WOMD) \citep{ettinger2021womd}. The Perceiver encoder \citep{jaegle2021perceiver} used in V-Max processes 280 input tokens---representing the ego vehicle, other agents, road geometry, traffic lights, and GPS route waypoints---through cross-attention layers, producing an explicit attention distribution over all scene elements at every timestep. By comparing three reward configurations (basic, minimal, complete) that share the same architecture and training data, we isolate the effect of reward design on learned attention patterns, evaluated across 50 real-world driving scenarios sampled from the WOMD validation set (3,676 timesteps for the complete model; 3,718 for the minimal model).

Our analysis yields a methodological contribution and two substantive findings, presented in order of evidential robustness:

\begin{enumerate}
    \item \textbf{Methodological foundation: within-episode correlation.} We establish that within-episode Spearman correlation $\rho_i$ between collision risk and agent-directed attention for each scenario $i$ is robustly positive across 31 high-variation scenarios for the complete-reward model (mean $\bar{\rho}{=}+0.291$ via Fisher z-transform (Eq.~\ref{eq:fisher}), 95\% confidence interval (CI) $[+0.125, +0.442]$, 80.6\% individually significant). In the 16 most risk-reactive scenarios ($\rho_i > 0.3$), agent attention increases by 77\% under threat ($\bar{\rho} = +0.522$, $p < 10^{-23}$, where $p$ denotes the statistical significance level). Crucially, na\"ive pooling across episodes yields $\rho{=}+0.088$---a $3.3\times$ underestimate due to between-scenario heterogeneity. This validated methodology enables the findings that follow.

    \item \textbf{Finding 1: Reward content directly shapes attention baselines.} GPS-path attention follows navigation reward content monotonically across reward configurations, with a $2.0\times$ ratio between the minimal and complete models (33.5\% vs.\ 16.4\%), and a $4.7\times$ range across all three models (from 7.1\% for basic to 33.5\% for minimal), computed over 50 scenarios each.

    \item \textbf{Finding 2: Continuous safety rewards create a vigilance prior.} Models trained with TTC-based proximity penalties maintain elevated agent attention even during collision-free phases: in 16 of 26 qualifying scenarios, the complete model maintains a mean of $+151\%$ more resting agent surveillance than the minimal model, with the gap present from the first timestep.
\end{enumerate}

We also identify a methodological implication for the XAI community: within-episode correlation with Fisher z-transform aggregation \citep{fisher1921probable} is the appropriate statistic for attention analysis across heterogeneous RL episodes, and per-scenario analysis reveals attentional heterogeneity that aggregate metrics conceal---including cases where identical scenarios produce opposite attention--risk correlations under different reward configurations.

\section{Related Work}
\label{sec:related}

\textbf{Reinforcement learning for autonomous driving.}
RL has gained traction in mid-to-end AD, where agents process structured scene representations rather than raw sensor data. V-Max \citep{charraut2025vmax} extends Waymax \citep{gulino2023waymax} with a JAX-based training pipeline, multiple encoders, and hierarchical reward structures that produce qualitatively different driving policies. Combining Imitation Learning (IL) with RL improves robustness in corner cases \citep{lu2023imitation}, while self-play yields highly robust policies \citep{cusumano2025robust}. The nuPlan challenge \citep{karnchanachari2024nuplan} showed that rule-based methods \citep{dauner2023pdm} still outperformed learning-based approaches, motivating further RL research. Incorporating RL objectives also helps mitigate the imitation gap in mid-to-end AD \citep{grislain2025igdrivsim}. We extend this line of work by showing that different reward configurations produce not only different policies but also qualitatively different internal attention patterns.

\textbf{Attention as explanation.}
Whether attention weights constitute valid explanations remains debated: \citet{jain2019attention} found them frequently uncorrelated with gradient-based importance, while \citet{wiegreffe2019attention} showed they can be informative when properly validated. In RL, perturbation-based saliency \citep{greydanus2018visualizing}, action-specific attribution \citep{gupta2020sarfa}, and exploratory analyses \citep{atrey2020exploratory} have been proposed, though their explanatory status remains contested. We sidestep the faithfulness debate entirely: rather than asking whether attention \textit{explains} decisions, we ask whether reward design \textit{shapes} attention patterns---a question about learned representations, not post-hoc explanation.

\textbf{Transformer architectures in driving.}
Transformer-based encoders are standard in motion forecasting and planning \citep{vaswani2017attention, nayakanti2023wayformer, shi2024mtr}, with hierarchical attention applied to AD imitation learning \citep{bronstein2022hierarchical}. The Perceiver \citep{jaegle2021perceiver}, used in V-Max's Latent-Query encoder (detailed in Section~\ref{sec:models}), processes heterogeneous inputs through cross-attention with learned latent queries, making attention weights directly interpretable as resource allocation across scene elements.

\textbf{Reward shaping.}
Reward design is a long-standing RL challenge \citep{sutton2018reinforcement}. In AD, hierarchical rewards produce qualitatively different policies \citep{charraut2025vmax}. We show that rewards change not only \textit{what the agent does} but also \textit{what it looks at}---and that these attentional effects are measurable and semantically coherent.

\section{Method}
\label{sec:method}

\subsection{Models and Reward Configurations}
\label{sec:models}

We study three SAC agents trained in V-Max \citep{charraut2025vmax} with the Latent-Query (LQ) encoder on the WOMD training set. The basic and minimal models were shared by 
\citet{charraut2025vmax} upon request and share the same 
architecture, training data, and initialisation seed (42) as the complete model, which we trained ourselves following the same V-Max setup.
\paragraph{Encoder architecture.}
The LQ encoder is a JAX re-implementation of the Perceiver architecture \citep{jaegle2021perceiver}. The Perceiver maps high-dimensional, heterogeneous inputs to a compact latent representation by means of \emph{cross-attention}: a fixed array of 16 learned latent vectors attends to the full 280-token input sequence via scaled dot-product attention \citep{vaswani2017attention}, compressing it into a representation that is then refined through 4 layers of self-attention (2 heads, head dimension $d_k = 16$). Because the latents---not the inputs---participate in self-attention, the computation cost is decoupled from input size. Crucially for our analysis, the cross-attention weights form an explicit, normalised distribution over all 280 input tokens at each layer and each timestep, providing a direct window into how the encoder allocates representational resources across scene elements.

\paragraph{Reward configurations.}
The three configurations form a strict hierarchy, each a superset of the previous, as shown in Table~\ref{tab:rewards}. Following \citet{charraut2025vmax}, the reward at each timestep $t$ is defined as follows:

\begin{align}
r^{\mathrm{basic}}_t \;=\;&
    -\mathcal{I}_{\mathrm{collision}}(t)
    - \mathcal{I}_{\mathrm{offroad}}(t)
    - \mathcal{I}_{\mathrm{red\text{-}light}}(t) \label{eq:r_basic}\\[4pt]
r^{\mathrm{minimal}}_t \;=\;&
    r^{\mathrm{basic}}_t
    - 0.2\cdot\mathcal{I}_{\mathrm{offroute}}(t)
    + 0.2\cdot\mathcal{I}_{\mathrm{progress}(t)>\mathrm{progress}(t-1)} \label{eq:r_minimal}\\[4pt]
r^{\mathrm{complete}}_t \;=\;&
    r^{\mathrm{minimal}}_t
    + 0.2\cdot\mathcal{I}_{\mathrm{comfort}}(t)
    - 0.1\cdot\mathcal{I}_{v(t)>v_{\mathrm{lim}}}
    - 0.2\cdot\max\!\left(0,\;1 - \frac{\mathrm{TTC}(t)}{\tau}\right) \label{eq:r_complete}
\end{align}

\noindent where $\mathcal{I}_{(\cdot)}$ denotes binary indicator functions and the final term of $r^{\mathrm{complete}}_t$ is a continuous shaped time-to-collision penalty. The \emph{basic} configuration penalises only safety violations. \emph{Minimal} adds a route-following incentive: a penalty for leaving the planned route and a reward for forward progress. \emph{Complete} additionally incorporates this TTC penalty (threshold $\tau{=}1.5$\,s; TTC computed as in Section~\ref{sec:risk}), a speed-limit compliance term, and a comfort incentive.

\begin{table}[ht]
    \caption{Reward configurations. Each successive configuration is a strict superset of the previous one, enabling controlled comparison of reward component effects. Accuracy (mean $\pm$ std over 3 seeds, WOMD validation set) 
is computed from checkpoints of the basic and minimal models 
provided by \citet{charraut2025vmax} and from our own 
trained complete model.}
    \label{tab:rewards}
    \begin{center}
    \begin{tabular}{llp{4cm}l}
        \toprule
        \textbf{Config.} & \textbf{Key addition} & \textbf{Reward terms} & \textbf{Accuracy} \\
        \midrule
        Basic    & Violation penalties only         & collision, offroad, red-light                               & $96.73 \pm 0.70\%$ \\
        Minimal  & Navigation incentive             & + off-route, progression                                   & $97.26 \pm 0.34\%$ \\
        Complete & Continuous proximity penalty     & + TTC ($\tau{=}1.5$\,s), speed, comfort                    & $97.44 \pm 0.37\%$ \\
        \bottomrule
    \end{tabular}
    \end{center}
\end{table}

All three models reach competitive accuracy on the WOMD validation set (Table~\ref{tab:rewards}). The complete model's best-seed checkpoint achieves 97.86\%---the top result in the V-Max benchmark \citep{charraut2025vmax}. Despite their near-identical accuracy, the three configurations differ substantially in their internal attention allocation, which is the subject of this paper.

\subsection{Attention Extraction}
\label{sec:attention}

As described in Section~\ref{sec:models}, the LQ encoder produces cross-attention weights over 280 input tokens at each timestep. Concretely, we extract these weights from the first cross-attention layer by reconstructing the attention matrix from the query ($Q$) and key ($K$) projections (with $d_k{=}16$, following \citealt{vaswani2017attention}):
\begin{equation}
    \mathbf{A} = \mathrm{softmax}\!\left(\frac{QK^\top}{\sqrt{d_k}}\right), \quad \mathbf{A} \in \mathbb{R}^{n_\mathrm{heads} \times n_\mathrm{queries} \times n_\mathrm{tokens}}.
\end{equation}
We average over heads and queries to obtain a single attention distribution over the 280 tokens, then aggregate into five semantic categories by summing over the corresponding token ranges.

\begin{table}[ht]
    \caption{Token structure of the LQ encoder input (280 tokens total).}
    \label{tab:tokens}
    \begin{center}
    \begin{tabular}{lcl}
        \toprule
        \textbf{Category} & \textbf{Token Indices (Count)} & \textbf{Content} \\
        \midrule
        Ego (SDC) & 0--4 \hspace{0.5em}(5) & Ego vehicle $\times$ 5 timesteps \\
        Other Agents & 5--44 \hspace{0.3em}(40) & 8 nearest agents $\times$ 5 timesteps \\
        Road Graph & 45--244 (200) & 200 sampled road points \\
        Traffic Lights & 245--269 (25) & 5 lights $\times$ 5 timesteps \\
        GPS Path & 270--279 (10) & 10 route waypoints \\
        \bottomrule
    \end{tabular}
    \end{center}
\end{table}

\subsection{Risk Metric}
\label{sec:risk}

We compute a continuous collision risk metric $\mathcal{R}$ from the simulation state at each timestep. Time-to-collision (TTC) is derived from the simulator state as follows: given the positions $\mathbf{p}_i$, velocities $v_i$, headings $\theta_i$, and bounding box dimensions of the ego vehicle and each surrounding agent $j$, the pairwise TTC is estimated under a constant-velocity assumption as the time until the two bounding boxes would first overlap:
\begin{equation}
    \mathrm{TTC}(j) = \frac{d_j - g_j}{\max(\dot{d}_j,\, \varepsilon)},
\end{equation}
where $d_j = \|\mathbf{p}_\mathrm{ego} - \mathbf{p}_j\|$ is the center-to-center distance, $g_j$ is the minimum safe separation accounting for bounding box extents, $\dot{d}_j = (v_\mathrm{ego} - v_j)\cos(\Delta\theta_j)$ is the closing speed along the line of approach, and $\varepsilon > 0$ prevents division by zero. The scene-level TTC is then $\min_j \mathrm{TTC}(j)$ over all valid agents within the observation window, following the V-Max evaluation protocol \citep{charraut2025vmax}. This is then mapped to a normalized collision risk score:
\begin{equation}
    \mathcal{R} = \mathrm{clip}\!\left(1 - \frac{\min_j\,\mathrm{TTC}(j)}{3.0},\ 0,\ 1\right).
\end{equation}
A value of $\mathcal{R}{=}0$ indicates no imminent risk ($\mathrm{TTC} \geq 3$s), and $\mathcal{R}{=}1$ indicates an imminent collision ($\mathrm{TTC}{=}0$).

\subsection{Within-Episode Correlation with Fisher Z-Transform}
\label{sec:fisher}

The central methodological choice in our analysis is to compute correlations \textit{within} each episode rather than pooling all timesteps across episodes.

\textbf{Why within-episode?} Na\"ive pooling conflates two sources of variation: (1)~the within-episode mechanism of interest---does attention shift when risk changes during a single drive?---and (2)~between-episode differences in baseline risk and attention levels, arising because urban scenarios have systematically different risk profiles and attention patterns than suburban ones. Episodes with near-constant risk (19 of 50 scenarios have $\mathrm{std}(\mathcal{R}) < 0.2$) contribute timesteps but no detectable signal, further diluting the pooled estimate.

We use Spearman rank correlation $\rho_i$ between $\mathcal{R}$ and agent attention for each scenario $i$ separately, filter to high-variation (HV) scenarios ($\mathrm{std} > 0.2$, $n{=}31$ for complete, $n{=}28$ for minimal), and aggregate using the Fisher z-transform \citep{fisher1921probable}:
\begin{equation}
\label{eq:fisher}
    z_i = \mathrm{arctanh}(\rho_i), \quad \bar{z} = \frac{1}{n}\sum_{i=1}^{n} z_i, \quad \bar{\rho} = \mathrm{tanh}(\bar{z}),
\end{equation}
where $\bar{\rho}$ is the aggregate correlation across scenarios. The z-transform maps $\rho_i$ values (which are compressed near $\pm1$) onto an approximately normal scale with known variance $\approx 1/(n_i - 3)$, where $n_i$ is the number of timesteps in scenario $i$, enabling proper averaging and confidence interval construction.

\section{Experiments}
\label{sec:experiments}

\subsection{Setup}

We evaluate the complete model on 50 scenarios sampled from the WOMD validation set (3,676 timesteps; 8 early terminations due to collisions or off-road events). The minimal model is evaluated on the same 50 WOMD validation scenarios (3,718 timesteps; no early terminations). The basic model is evaluated on 3 overlapping scenarios (157 timesteps; 2 early terminations due to crashes), providing a reference point for the navigation-free regime. These 50 scenarios constitute the fixed analysis set used throughout the paper and are all drawn from the WOMD validation split. At each timestep, we record the five category-level attention fractions, collision risk $\mathcal{R}$, ego actions (acceleration, steering), and scene metadata (number of valid agents, collision/off-road flags). All simulations run at 10~Hz for up to 80~timesteps (8 seconds of driving after a 1-second log-replay warm-up).

\subsection{Establishing the Methodology: Within-Episode Attention Tracks Collision Risk}
\label{sec:methodology_validation}

Across 31 high-variation scenarios, within-episode Spearman correlation $\rho_i$ between $\mathcal{R}$ and agent attention is consistently positive (Table~\ref{tab:correlations}). We present this result first not because it is the most conceptually interesting finding, but because it establishes---on the full 50-scenario dataset---that within-episode attention analysis is a valid and reliable methodology before the cross-model comparisons that follow.
\begin{table}[ht]
    \caption{Within-episode correlations (complete-reward model, 31 HV scenarios). $\bar{\rho}$ and 95\% CI computed via Fisher z-transform (Eq.~\ref{eq:fisher}). ``Sig.\ \%'' is the fraction of individual scenario correlations $\rho_i$ that are statistically significant at $p < 0.05$.}
    \label{tab:correlations}
    \begin{center}
    \begin{tabular}{lccc}
        \toprule
        \textbf{Pair} & $\bar{\rho}$ & \textbf{95\% CI} & \textbf{Sig.\ \%} \\
        \midrule
        $\mathcal{R} \times \text{attn\_agents}$     & $+0.291$ & $[+0.125, +0.442]$ & 80.6\% \\
        $\mathcal{R} \times \text{attn\_roadgraph}$  & $-0.148$ & $[-0.287, -0.003]$ & 67.7\% \\
        \bottomrule
    \end{tabular}
    \end{center}
\end{table}

The confidence interval for agent attention is entirely above zero. The road graph shows a complementary negative correlation: attention shifts \textit{from} static road geometry \textit{toward} dynamic agents when collision risk rises. This bidirectional reallocation is semantically coherent---the model prioritizes dynamic threats over static infrastructure under danger.

\textbf{The pooling confound.} Na\"ive pooling across all 3,676 timesteps gives $\rho{=}+0.088$ ($p < 0.0001$)---a $3.3\times$ underestimate. The gap arises because 19 low-variation scenarios contribute timesteps but no detectable signal, and because between-scenario baseline differences in both risk and attention create a confound. This is our core methodological finding: pooled analysis is misleading for heterogeneous RL episodes.

\textbf{Agent-count confound check.} A concern is that higher collision risk simply coincides with more nearby agents, mechanically increasing agent attention. We compute partial correlation controlling for the number of valid agents: raw $\rho{=}+0.262$, partial $\rho{=}+0.247$ ($\Delta{=}-0.014$). The confound is negligible.

\begin{figure}[t]
    \centering
    \includegraphics[width=\linewidth]{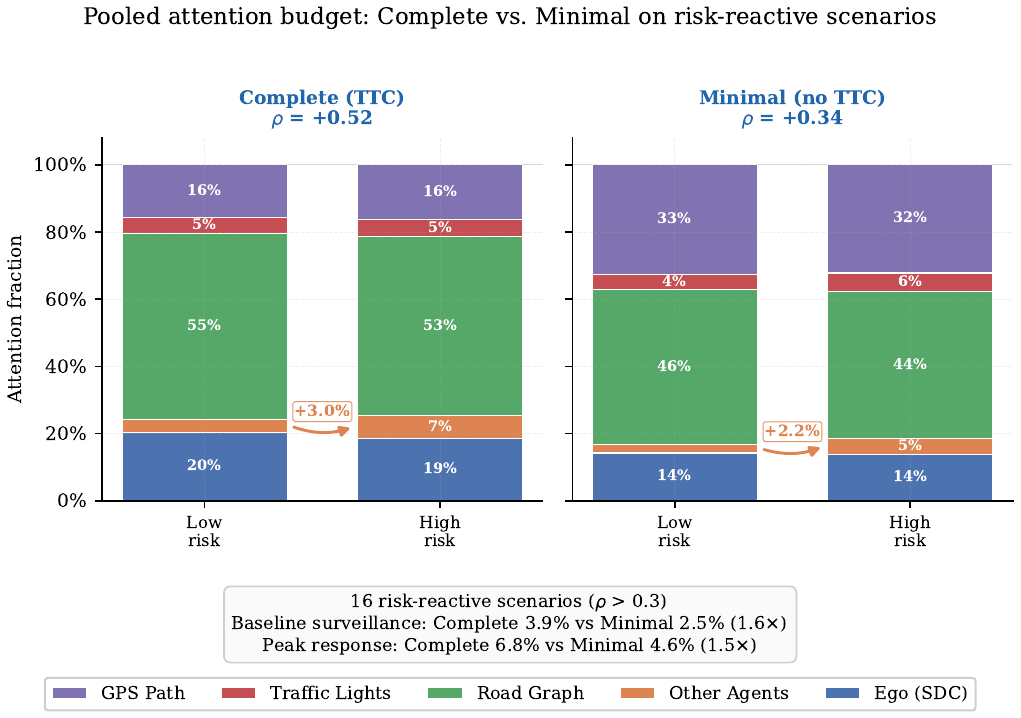}
    \caption{Pooled attention budget under low risk ($\mathcal{R} < 0.2$) vs.\ high risk ($\mathcal{R} > 0.7$) in the 16 risk-reactive scenarios ($\rho > 0.3$). Both models reallocate attention from Road Graph toward Other Agents under threat. The complete (TTC) model maintains 1.6$\times$ higher baseline agent surveillance (3.9\% vs.\ 2.5\%) and 1.5$\times$ higher peak response (6.8\% vs.\ 4.6\%) relative to the minimal model. Road Graph attention is dominant in both conditions because it is represented by 200 of 280 tokens; the agent shifts, while numerically small in absolute terms, represent substantial proportional increases for a category with limited token budget.}
    \label{fig:budget}
\end{figure}

\textbf{Budget reallocation under threat.} In the 16 risk-reactive scenarios (within-episode $\rho > 0.3$, representing 57\% of all high-variation episodes), the complete model increases agent attention by $+76.7\%$ from low to high risk ($p = 1.24 \times 10^{-23}$), with a Fisher z-transformed mean $\rho = +0.522$ (95\% CI $[+0.456, +0.582]$). Although the Other Agents category occupies only 40 of 280 tokens (14\% of the input), it captures a mean attention fraction of 3.9\% at low risk and 6.8\% at high risk in the complete model. These absolute fractions are modest, but because the attention budget is closed and road geometry dominates the token count, even a 3-percentage-point reallocation away from the largest category (Road Graph) represents a proportionally large increase for a small-token category. The minimal model shows a similar reactive pattern but from a lower baseline: 2.5\% rising to 4.6\% ($+89.1\%$, $p = 4.70 \times 10^{-15}$), consistent with a weaker surveillance prior (Figure~\ref{fig:budget}).

\textbf{Counter-examples.} A minority of scenarios (5 of 28 HV episodes in the complete model) show reversed correlations ($\rho < 0$), reflecting genuine attentional heterogeneity in which the agent increases attention to road geometry during high-risk phases; we discuss these in the supplementary material.

\subsection{Finding 1: Reward Content Shapes Attention Baselines}
\label{sec:finding1}

\begin{figure}[t]
    \centering
    \includegraphics[width=\linewidth]{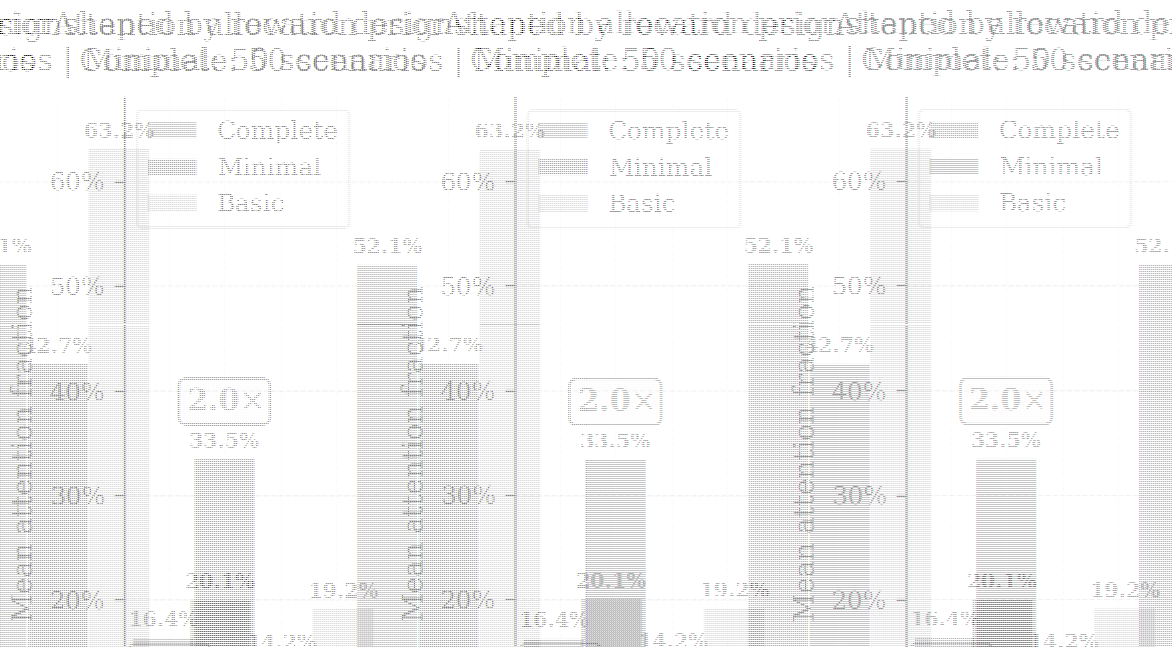}
    \caption{Attention allocation prior shaped by reward design, averaged over 50 scenarios for the complete and minimal models and 3 scenarios for the basic model. GPS-path attention follows navigation reward content: the minimal model (33.5\%) allocates $2.0\times$ more than the complete model (16.4\%), which allocates $2.3\times$ more than the basic model (7.1\%). The basic model, lacking navigation rewards, compensates with elevated road graph attention (63.2\%). The complete model allocates $1.3\times$ more attention to other agents than the minimal model (5.6\% vs.\ 4.2\%), consistent with the vigilance prior from TTC training.}
    \label{fig:gps_gradient}
\end{figure}

Comparing episode-averaged attention across all 50 scenarios reveals that attention baselines directly track reward content (Figure~\ref{fig:gps_gradient}).

\textbf{GPS attention gradient.} GPS-path attention follows navigation reward content monotonically across configurations. The minimal model, which receives navigation incentives but no TTC penalty, allocates 33.5\% of its attentional budget to GPS route tokens. Adding TTC penalties (complete) redistributes this budget toward agents and road geometry, bringing GPS attention down to 16.4\%---a $2.0\times$ reduction. Removing navigation rewards entirely (basic) renders GPS nearly irrelevant at 7.1\%, a $4.7\times$ reduction relative to the minimal model. A model with no route incentive has no reason to monitor route waypoints, and this is exactly what the attention patterns reflect.

\begin{table}[h!]
    \caption{Vigilance gap: calm-phase ($\mathcal{R} < 0.2$) agent attention for the complete (TTC) and minimal (no TTC) models. Top three scenarios shown alongside the cross-scenario mean.}
    \label{tab:vigilance}
    \begin{center}
    \begin{tabular}{lccc}
        \toprule
        \textbf{Scenario} & \textbf{Complete (TTC)} & \textbf{Minimal (no TTC)} & \textbf{Gap} \\
        \midrule
        s002 & 0.144 & 0.077 & $+86\%$ \\
        s016 & 0.042 & 0.006 & $+564\%$ \\
        s021 & 0.049 & 0.022 & $+129\%$ \\
        \midrule
        \textbf{Mean (16 scen.)} & --- & --- & $\mathbf{+151\%}$ \\
        \bottomrule
    \end{tabular}
    \end{center}
\end{table}

\textbf{Road graph compensation.} The basic model, deprived of any navigation signal, compensates by dramatically elevating road graph attention to 63.2\%---substantially higher than the complete (52.1\%) and minimal (42.7\%) models. Without a GPS path to follow, the agent relies more heavily on local road geometry for navigation decisions.

\textbf{Agent attention baseline.} The complete model allocates $1.3\times$ more baseline attention to other agents than the minimal model (5.6\% vs.\ 4.2\%), consistent with the vigilance prior documented in \hyperlink{find:2}{Finding~2}  below. Although Other Agents account for only 14\% of input tokens, this 1.4-percentage-point difference is meaningful within the closed attentional budget: it represents a systematically higher resource allocation to dynamic scene elements at the expense of static geometry.

\subsection{Finding 2: TTC Reward Creates a Vigilance Prior}
\label{sec:finding2}

The most conceptually important finding concerns the \textit{resting} level of agent attention during collision-free phases. The vigilance prior is confirmed at scale across 50 scenarios. In 16 of 26 qualifying scenarios (those with sufficient calm-phase timesteps for both models), the complete model maintains higher agent attention during collision-free phases ($\mathcal{R} < 0.2$), with a mean gap of $+151\%$ (Table~\ref{tab:vigilance}).

The gap is present from $t{=}0$---before any danger appears---and persists through calm phases. This is not a reactive effect. The TTC penalty is a \textit{continuous} reward signal that fires whenever TTC drops below 1.5 seconds, training persistent agent surveillance: the model learns that other agents are permanently relevant, even in currently safe situations. We term this a \textit{vigilance prior}: a learned resting attentional posture shaped by reward design.

\begin{figure}[t]
    \centering
    \includegraphics[width=\linewidth]{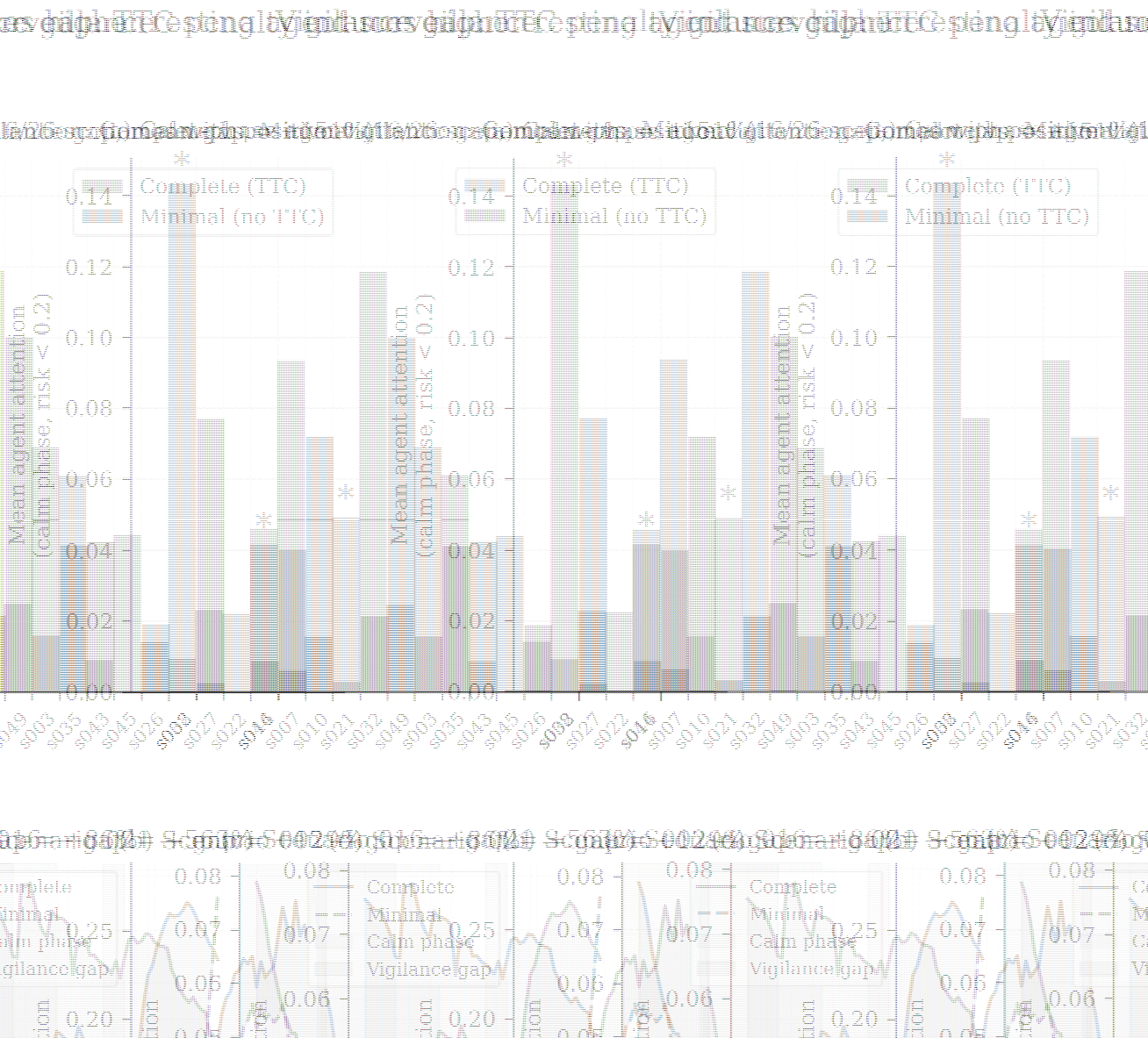}
    \caption{Vigilance gap across 50 scenarios. \textbf{(a)} Calm-phase ($\mathcal{R} < 0.2$) agent attention for both models across the 16 qualifying scenarios. In 16 of 26 qualifying scenarios (62\%), the TTC-penalized model maintains higher agent attention during safe phases (mean gap $= +151\%$). \textbf{(b--d)} Timeseries for three representative scenarios showing the complete model (solid) consistently above the minimal model (dashed) during calm phases (green shading), with the vigilance gap (blue fill) present from $t{=}0$.}
    \label{fig:vigilance}
\end{figure}

\begin{figure}[h!]
    \centering
    \includegraphics[width=\linewidth]{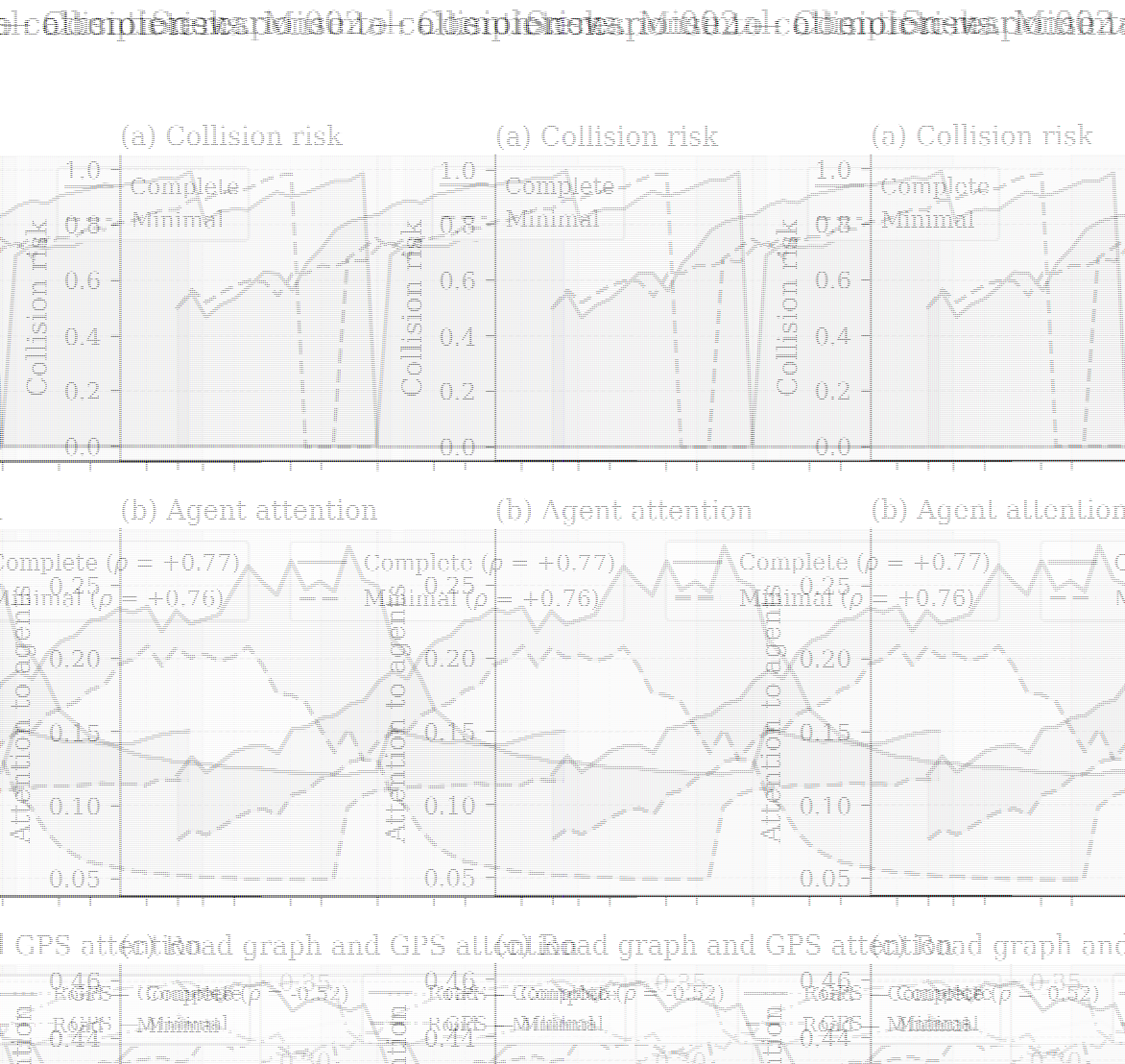}
    \caption{Scenario 002: Complete vs.\ Minimal attention response to collision risk. \textbf{(a)} Collision risk $\mathcal{R}$ over 80 timesteps, with danger (red) and calm (green) phases highlighted. \textbf{(b)} Agent attention: both models track risk positively ($\rho = +0.77$ and $\rho = +0.76$, respectively), but the complete model maintains a higher baseline throughout (vigilance prior). \textbf{(c)} Road graph and GPS attention: the complete model shows lower GPS attention (16\% vs.\ 33\%) and a stronger negative road-graph correlation under threat, reflecting the attentional budget reallocation toward agents.}
    \label{fig:timeseries}
\end{figure}

\textbf{Caveats.} The complete and minimal models face partially different risk experiences on the same scenarios because their different policies lead to different ego trajectories. The vigilance gap is therefore a property of each model's learned attentional posture rather than a controlled comparison of responses to identical stimuli. Furthermore, in 10 of 26 qualifying scenarios the gap is absent or reversed, indicating that vigilance effects interact with scenario-specific geometry that warrants further investigation.

\textbf{Reward-driven attention reversal.} The most striking evidence that reward design shapes attention strategy comes from scenarios where the two models exhibit \textit{opposite} attention--risk correlations. Figure~\ref{fig:timeseries} illustrates scenario s002, where both models show strong positive correlation: agent attention rises during high-risk phases for both the complete ($\rho_i = +0.77$) and minimal ($\rho_i = +0.76$) models, yet the complete model maintains systematically higher agent attention throughout the episode. More extreme reversals occur in 5 of the 16 strong-$\rho_i$ scenarios, where the complete-reward model shows positive correlation and the minimal-reward model shows negative correlation on the same episode. Since both models share the same architecture and training 
data, and were trained with the same initialisation seed (42), 
this qualitative reversal cannot be attributed to any of 
these factors. The reward function is the only degree of freedom, and because attention weights form a closed budget, different reward signals induce different allocation strategies: the TTC-trained model directs attention toward dynamic agents under threat, while the minimal-reward model redistributes toward road geometry to plan avoidance manoeuvres, producing opposite correlation signs from the same attentional budget constraint.

\section{Discussion}
\label{sec:discussion}

\textbf{Reward shapes representations, not just behavior.} Our results demonstrate that reward design in RL does not merely change what the agent \textit{does}---it changes what the agent \textit{looks at}. GPS attention tracks navigation reward content with a $2.0\times$ ratio between the minimal and complete models (and a $4.7\times$ range across all three), and TTC-based proximity penalties produce a measurable vigilance prior. This suggests that attention analysis can serve as a practical diagnostic: if a reward function is intended to promote safety awareness, one can verify that the encoder allocates elevated attention to dynamic agents, and vice versa.

\textbf{Within-episode analysis is essential.} The $3.3\times$ gap between pooled ($\rho{=}+0.088$) and within-episode ($\rho{=}+0.291$) estimates has broad methodological implications. Analyses that pool observations across heterogeneous RL episodes risk underestimating---or entirely missing---meaningful attention--environment relationships. We recommend within-episode correlation with Fisher z-transform aggregation as a standard practice whenever attention is analyzed across RL trajectories.

\textbf{The proportional significance of small token categories.} When interpreting attention fractions, the token count asymmetry in the input must be kept in mind. Road graph tokens (200 of 280, 71\%) dominate the input, so road graph attention will naturally be the largest fraction. Conversely, Other Agents (40 tokens, 14\%) and GPS Path (10 tokens, 4\%) are structurally smaller, so a shift of a few percentage points in agent or GPS attention represents a much larger proportional reallocation within those categories and should be interpreted as a meaningful behavioral signal.

\begin{figure}[t]
    \centering
    \includegraphics[width=\linewidth]{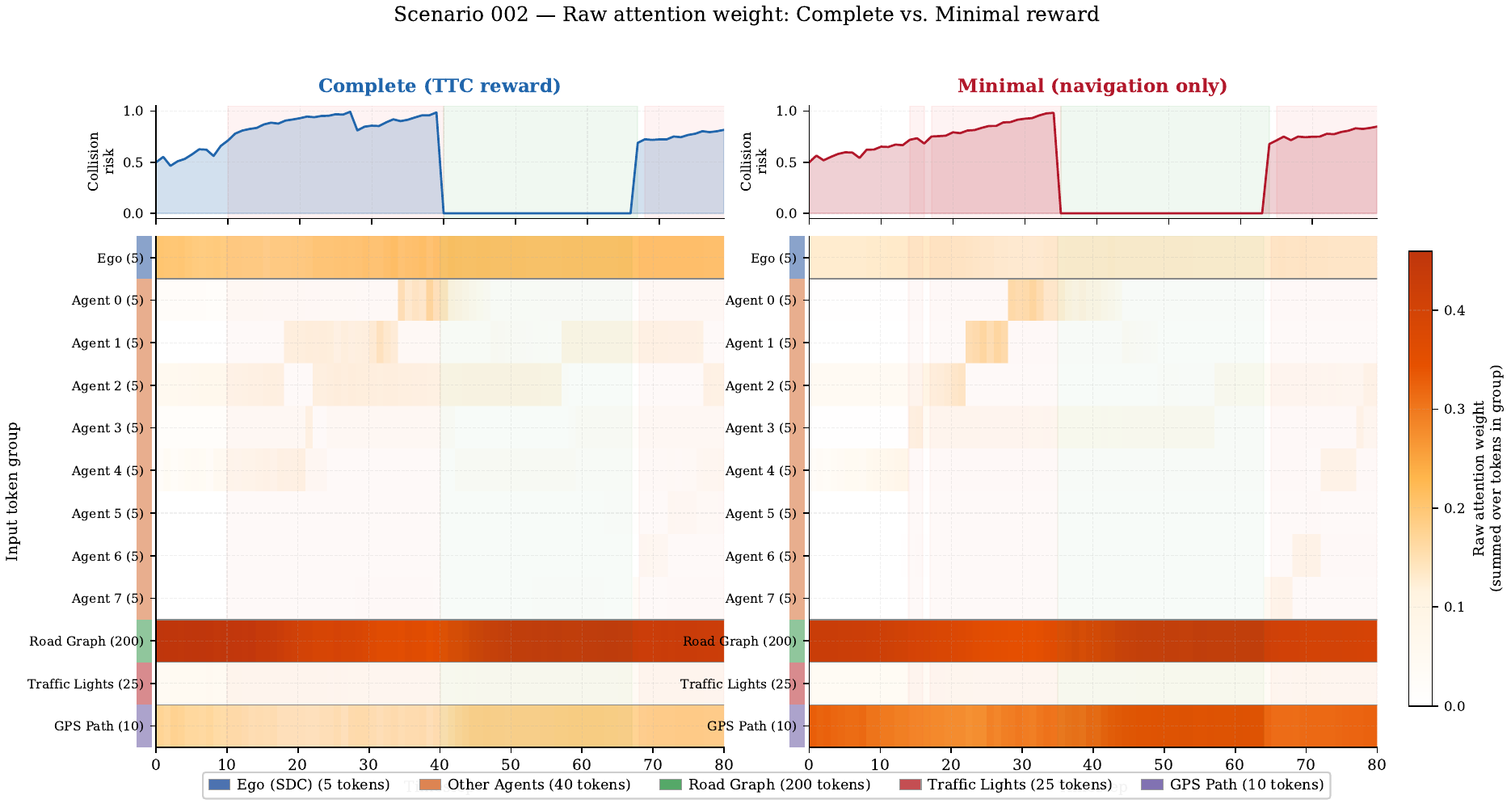}
    \caption{Raw attention map for scenario s002 (complete vs.\ minimal model). Each row shows the aggregated attention fraction for one category over the episode. Road Graph (200 tokens) dominates with ${\sim}52\%$ of the total budget, obscuring dynamics in smaller categories. The minimal model's elevated GPS-path attention (bottom row) is visible, but agent-level modulation under risk is largely masked by the Road Graph's dominance.}
    \label{fig:attn_raw}
\end{figure}

\begin{figure}[h!]
    \centering
    \includegraphics[width=\linewidth]{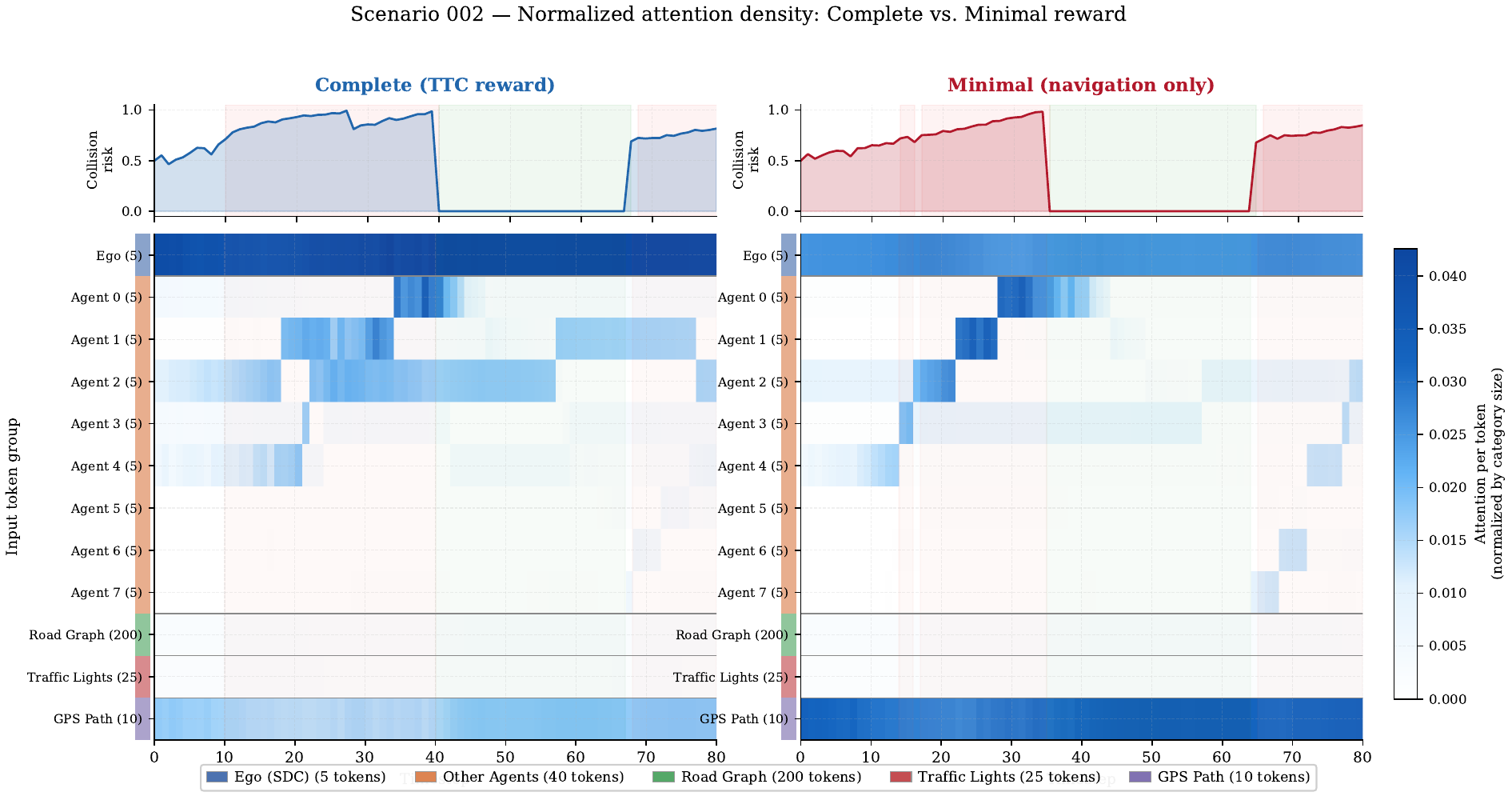}
    \caption{Token-normalized attention map for scenario s002 (complete vs.\ minimal model). Each category's attention is divided by its token count (Table~\ref{tab:tokens}), revealing attention density per token. Ego has the highest per-token density in both models. During danger phases, individual agent tokens in the complete model receive 5--8$\times$ more attention per token than road graph tokens, confirming selective threat monitoring. The GPS density gap between models is now directly comparable to agent density: the minimal model allocates similar per-token attention to GPS and agents, while the complete model's TTC penalty redirects budget toward dynamic threats at the expense of GPS.}
    \label{fig:attn_norm}
\end{figure}

To make this asymmetry explicit, we define the \textit{token-normalized attention} for each category $c$ as $\hat{a}_c = a_c / w_c$, where $a_c$ is the raw attention fraction and $w_c$ is the category's token weight. After normalization, GPS Path ($\hat{a}_c = 4.56$) and Ego ($\hat{a}_c = 2.78$) emerge as the most attended categories \textit{per token}, while Road Graph ($\hat{a}_c = 0.73$)---despite commanding over half the raw budget---receives below-proportional attention. This confirms that the reward-conditioned effects we report (GPS gradient, vigilance prior) are not artifacts of category size but reflect genuine representational priorities. Figures~\ref{fig:attn_raw} and~\ref{fig:attn_norm} visualize this on scenario s002: the raw map (Figure~\ref{fig:attn_raw}) shows Road Graph dominating at ${\sim}52\%$, masking dynamics in smaller categories, while normalization (Figure~\ref{fig:attn_norm}) reveals the per-token density patterns discussed above.

Dividing each category's attention by its token count (Table~\ref{tab:tokens}) yields the normalized attention map (Figure~\ref{fig:attn_norm}), revealing attention \textit{density per token}. This normalization exposes three patterns hidden in the raw view: Ego has the highest per-token density in both models; individual agent tokens receive 5--8$\times$ more attention than road graph tokens during danger phases in the complete model, confirming selective threat monitoring; and GPS density in the minimal model matches agent density, while in the complete model it drops below agents---the TTC penalty redirects budget toward dynamic threats.

\textbf{Counter-examples are informative.} The cases where the complete and minimal models exhibit opposite attention--risk correlations on the same scenario are the most analytically valuable results. They demonstrate that reward design can qualitatively reverse attentional strategy, not merely modulate its magnitude---a stronger claim than simple amplitude differences.

\textbf{Attention as representation, not explanation.} We deliberately frame our analysis as studying \textit{reward-conditioned representations} rather than claiming attention \textit{explains} decisions. Following the distinction drawn by \citet{jain2019attention} and \citet{wiegreffe2019attention}, attention weights may not faithfully reflect the causal chain from input to output. Our claim is narrower and distinct: reward design shapes the \textit{distribution} of attention across input categories in a predictable and semantically coherent manner.

\section{Conclusion}
\label{sec:conclusion}

We have shown that reward design in RL-trained autonomous driving agents predictably shapes the Perceiver encoder's cross-attention allocation. Using within-episode correlation---validated as a reliable methodology across 50 WOMD validation scenarios for both the complete and minimal models---we demonstrate that GPS attention tracks navigation reward content monotonically and that continuous proximity penalties produce a learned vigilance prior. The $3.3\times$ gap between pooled and within-episode correlation estimates highlights a methodological pitfall relevant to the broader XAI community. Future work should investigate the structural causes of attention reversal scenarios, extend the analysis to other encoder architectures (Wayformer, MTR), and test whether reward-conditioned attention patterns generalize across driving domains and observation modalities.

\subsubsection*{Broader Impact Statement}
\label{sec:broaderImpact}
This work contributes to the transparency of RL-based autonomous driving agents by providing tools to verify that reward functions produce the intended attentional behavior. Improved understanding of RL agent internals may accelerate deployment of autonomous systems; however, attention-based analysis should complement---not substitute for---rigorous safety testing and validation.

\subsubsection*{Acknowledgments}
This work was carried out partially using the AI Datacenter of the National School of Artificial Intelligence, funded under grant number E049 24 0117 by the Algerian Ministry of Higher Education and Scientific Research.


\bibliography{main}
\bibliographystyle{elsarticle-harv}


\newpage
\appendix

\section*{Supplementary A: Scenario Selection and Filtering}
\label{sec:app_scenarios}

Of 50 scenarios evaluated with the complete model, 31 pass the high-variation filter ($\mathrm{std}(\mathcal{R}) > 0.2$); for the minimal model, 28 of 50 scenarios pass the same filter. Effective scenarios for within-episode correlation analysis require risk variability with both calm and dangerous phases. We found that raw event count (total detected collision/off-road/violation events) is not a reliable proxy for analytical value: the scenario with the highest event count (48) had near-constant maximum risk, rendering it uninformative for correlation analysis. We recommend selecting scenarios based on risk standard deviation and the presence of multiple risk cycles.

\section*{Supplementary B: Cross-Model Risk-Reactivity}
\label{sec:app_crossmodel}

On the overlapping scenario s002, all three models show strong positive within-episode correlations between $\mathcal{R}$ and agent attention, indicating risk-reactive attention is a general property of collision-avoidance training:

\begin{table}[ht]
    \caption{Within-episode $\rho(\mathcal{R},\, \text{attn\_agents})$ on s002 across reward configurations. $^{**}$: $p < 0.01$.}
    \begin{center}
    \begin{tabular}{lcc}
        \toprule
        \textbf{Model} & $\rho$ & \textbf{Episode length} \\
        \midrule
        Complete (TTC) & $+0.769^{**}$ & 80 timesteps \\
        Minimal (no TTC) & $+0.765^{**}$ & 80 timesteps \\
        Basic (violations only) & $+0.990^{**}$ & 39 timesteps (early termination) \\
        \bottomrule
    \end{tabular}
    \end{center}
\end{table}

What differs between models is the \textit{baseline level} from which the reactive response occurs and the \textit{survival rate}: the basic model crashes in 2 of 3 evaluated scenarios.

\textbf{Road graph compensation.} Without GPS route guidance, the basic model compensates by allocating substantially more attention to road geometry (0.476 vs.\ 0.419 for complete and 0.402 for minimal on s002). During braking, 72\% of the basic model's attention budget goes to road graph, compared to approximately 42\% for the complete model.

\section*{Supplementary C: Counter-Example Scenarios}
\label{sec:app_counterexamples}

Five of the 16 strong-$\rho$ scenarios in the complete model ($\rho > 0.5$) show reversed correlations in the minimal model (negative $\rho$ on the same scenario). In scenario s023, the complete model shows $\rho = +0.61$ while the minimal model shows $\rho = -0.39$: agent attention rises under threat for the complete model and falls under threat for the minimal model. Similarly, s000 shows $\rho = +0.62$ vs.\ $-0.08$, and s027 shows $\rho = +0.30$ vs.\ $-0.46$. The most plausible explanation is that without a TTC penalty, the minimal model responds to increasing risk by attending more to road geometry (to plan avoidance trajectories), while the TTC-trained complete model has already internalized a surveillance habit that directs it toward agents. These cases demonstrate that reward design is the single degree of freedom separating opposite attentional strategies on identical inputs.

\section*{Supplementary D: Attention Budget Verification}
The softmax-normalized attention weights sum to exactly 1.0 at every timestep across all 3,676 observations. We verified this programmatically with a tolerance of $10^{-6}$ and found zero violations. This invariant confirms that our 5-category aggregation correctly represents a closed budget system.

\section*{Supplementary E: Entropy Analysis Details}
Shannon entropy is computed over the five attention categories: $H_t = -\sum_{c=1}^{5} p_c \log_2 p_c$, where $p_c$ is the attention fraction for category $c$. Theoretical range is $[0, \log_2 5] = [0, 2.322]$ bits. Observed range across all timesteps is $[0.91, 2.19]$ bits.

Within-episode correlation between collision risk and entropy:
\begin{itemize}
    \item Mean across HV scenarios: $\rho = +0.199$
    \item Calm-phase mean entropy: $H = 1.770$ bits
    \item Danger-phase mean entropy: $H = 1.803$ bits ($p < 0.0001$, Mann-Whitney U)
\end{itemize}

The positive direction (entropy \textit{increases} with risk) was initially unexpected, as one might predict that the model would narrow focus under threat. Instead, the increase reflects the bidirectional reallocation documented in Finding~1: attention shifts from the dominant road-graph category toward the smaller agent category, producing a more uniform distribution and thus higher entropy.

\end{document}